\documentclass[%
 reprint,
 amsmath,amssymb,
 aps,
]{revtex4-1}

\usepackage{graphicx}
\usepackage{dcolumn}
\usepackage{bm}
\usepackage[ruled,vlined]{algorithm2e}
\usepackage{subfigure}

\begin{document}

\preprint{APS/123-QED}

\title{Using Deep LSD to build operators in GANs latent space\\
with meaning in real space}

\author{J. Quetzalc\'oatl Toledo-Mar\'in}
    \email{j.toledo.mx@gmail.com}
\author{James A. Glazier}%
\affiliation{%
 Biocomplexity Institute, Indiana University,\\
  Bloomington, IN 47408, USA\\
  Luddy School of Informatics,\\
  Computing and Engineering, IN 47408, USA
}%




\date{\today}

\begin{abstract}
Generative models rely on the key idea that data can be represented in terms of latent variables which are uncorrelated by definition. Lack of correlation is important because it suggests that the latent space manifold is simpler to understand and manipulate. Generative models are widely used in deep learning, \textit{e.g.}, variational autoencoders (VAEs) and generative adversarial networks (GANs). Here we propose a method to build a set of linearly independent vectors in the latent space of a GANs, which we call quasi-eigenvectors. These quasi-eigenvectors have two key properties: i) They span all the latent space, ii) A set of these quasi-eigenvectors map to each of the labeled features one-on-one. We show that in the case of the MNIST, while the number of dimensions in latent space is large by construction, $98\%$ of the data in real space map to a sub-domain of latent space of dimensionality equal to the number of labels. We then show how the quasi-eigenvalues can be used for Latent Spectral Decomposition (\textit{LSD}), which has applications in denoising images and for performing matrix operations in latent space that map to feature transformations in real space. We show how this method provides insight into the latent space topology. The key point is that the set of quasi-eigenvectors form a basis set in latent space and each direction corresponds to a feature in real space.
\end{abstract}

\maketitle


\section{Introduction}
 \textit{Generative models} (GMs) are a class of Machine Learning (ML) model which excel in a wide variety of tasks \cite{noe2020machine}. The optimization of a GM finds a function $\mathcal{G}$ that maps a set of \textit{latent variables} in \textit{latent space} to a set of variables in real space representing the data of interest (\textit{e.g.}, images, music, video, \textit{etc}.), \textit{i.e.} $\mathcal{G}: \Re^M \rightarrow \Re^d$ where $d > M \gg 1$. When building a GM, we first define the support of the latent variables, then obtain the function $\mathcal{G}$ by optimizing a loss function. Loss function choice depends on application, \textit{e.g.}, maximum log-likelihood is common in Bayesian statistics \cite{mackay2003information}, Kullback–Leibler divergence  is common for variational autoencoders (VAEs) \cite{kingma2017variational} and the Jensen-Shannon entropy and the Wasserstein distance are common  with generative adversarial networks (GANs) \cite{goodfellow2016nips}.
Latent variables have a simple distribution, often a separable distribution (\textit{i.e.,} $P(\lbrace z_i \rbrace_{i=1}^M) = \prod_{i=1}^M P(z_i)$). Thus, when we fit a latent variable model to a data set, we are finding a description of the data in terms of "independent components" \cite{mackay2003information}.
Often the latent representation of data lives in a simpler manifold than the original data while preserving \textit{relevant} information.
For instance, Ref. \cite{gardner2006sparse} proposes a time-frequency representation of a signal that allows the reconstruction of the original signal, which relies in what they define as "\textit{consensus}". Their proposed method generates sharp representations for complex signals.

Trained deep neural network can function as surrogate propagators  for time evolution of physical systems  \cite{noe2020machine}. While the latent variables are constructed to be independent identically distributed (i.i.d.) random variables, training \textit{entangles} these latent variables. Latent variable disentanglement is an  active area of research employing a wide variety of methods. 
For instance, in Ref. \cite{peebles2020hessian}, the authors train a GAN  including  the generator's Hessian as a regularizer in the loss function,  leading, in optimum conditions, to linearly independent latent variables, where each latent variable independently controls the strength of a single feature. 
Ref. \cite{razavi2019generating} constructs a set of \textit{quantized} vectors in the latent space using a VAE, known as \textit{vector quantized variational autoencoder} (VQ-VAE). Each \textit{quantized} vector highlights a specific feature of the data set. This approach has been used in OpenAI's jukebox \cite{dhariwal2020jukebox}. A major drawback of these approaches is the lack of freedom in relating specific features in real space with specific latent space directions. This can be overcome by \textit{conditionalizing} the generative model \cite{perarnau2016invertible}. However, conditionalization can reduce the latent space \textit{smoothness} and \textit{interpolation capacity}, since the condition is usually enforced by means of discrete vectors as opposed to a continuous random latent vector. 

Here we propose a method to relate a specific chosen labeled feature with specific directions in latent space such that these directions are linearly independent. Having a set of linearly-independent latent vectors associated with specific labeled features allows us to define operators that act on latent space (\textit{e.g.} a rotation matrix) and correspond to feature transformations in real space. For instance, suppose a given data set in real space corresponds to the states of a molecular dynamic simulation, \textit{i.e.}, $|x_i \rangle \rightarrow |x(t_i) \rangle$ and suppose $|x(t_i) \rangle = \mathcal{G} | z_i \rangle$ and $|x(t_i + \Delta t) \rangle = \mathcal{G} | z_j \rangle$. How can we construct an operator in latent space, $\mathcal{O}_{\Delta t}$, such that $|z_j \rangle = \mathcal{O}_{\Delta t} | z_i \rangle$?. For construction to be possible, the operator $\mathcal{G}$ must be \textit{locally} linear. Furthermore, in order to build the operator $\mathcal{O}$, we need a basis that \textbf{spans} latent space. While linearity these might seem counterintuitive given  how NNs work, growing evidence suggests such linearity in practice. For instance, there is an ongoing debate on how deep should a NN be to perform a specific task. Moreover, it has been proposed the equivalence between deep NNs and shallow wide NNs \cite{bahri2020statistical}. For at least one image-related GAN, simple vector arithmetic in latent space leads to feature transformations in real space (\textit{e.g.}, removal of sunglasses, change in hair color, gender, \textit{etc}.) \cite{radford2015unsupervised}. However, we still do not understand how specific features in real space map to latent space and how are these features arranged in latent space (\textit{latent space topology}) or why some GANs  behave like linear operators.
The latent representation of data with a given labeled feature forms a cluster. However, the tools employed to show this clustering effect quite often consist in a dimensional reduction \textit{e.g.}, t-SNE \cite{maaten2008visualizing} collapses the latent representation into two or three dimensions. Other methods include principal component analysis, latent component analysis and important component analysis \cite{jolliffe1986principal, mackay2003information, muthen2004latent}. Our method does not collapse or reduce the latent space, allowing us to inspect latent space topology by spanning  all latent space directions. We demonstrate the method by applying it to MNIST.

\begin{figure}[hbtp]
\centering
\includegraphics[width=3.2in]{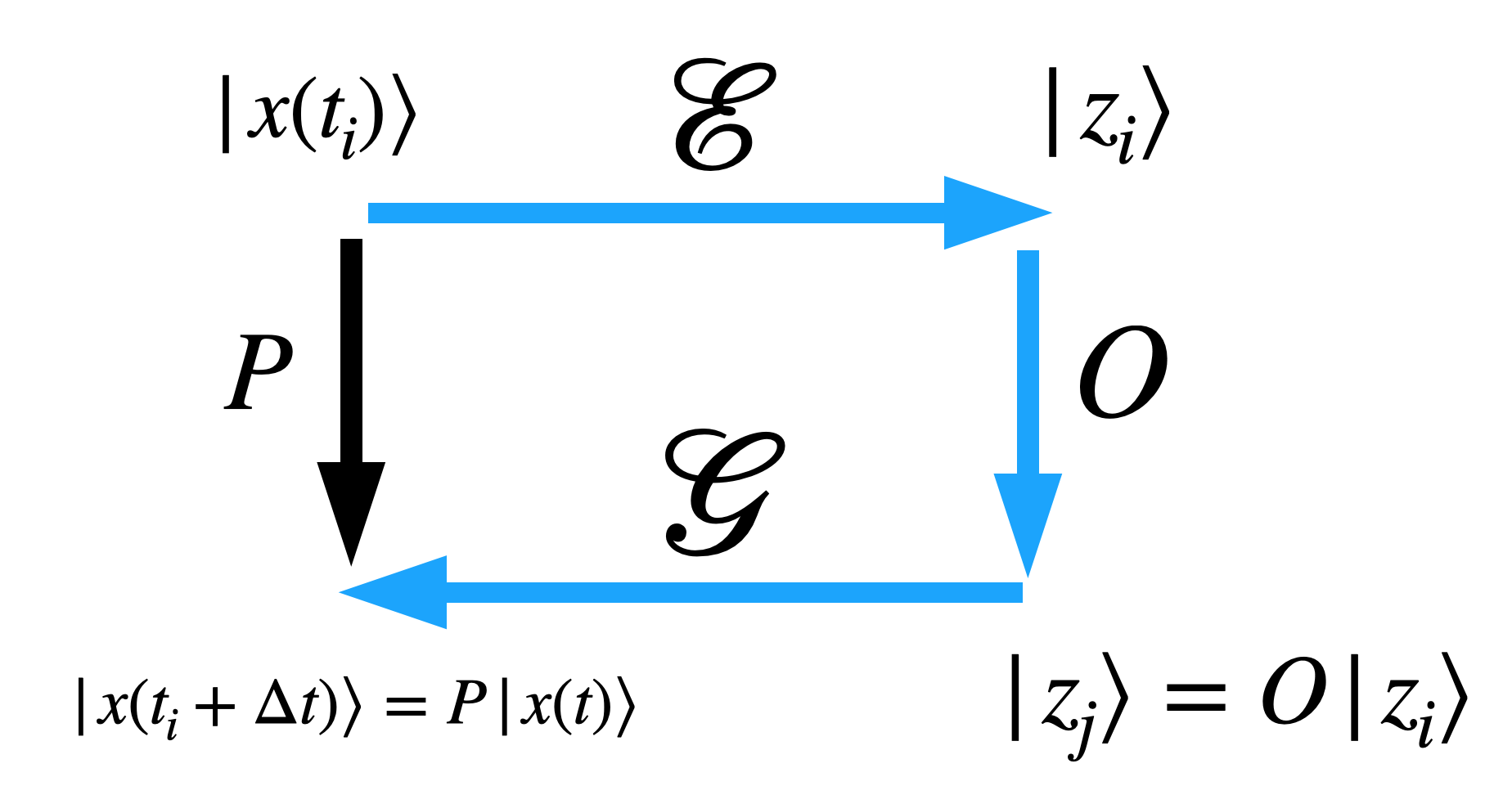}
\caption{ Schematic of spaces and operators. $P$ is an operator in real space that evolves the state $|x(t_i) \rangle$ to $|x(t_i + \Delta t) \rangle$. $\mathcal{E}$ is an Encoder and $\mathcal{G}$ is the Generator that maps latent variables to real space. $O$ is an operator in latent space. The black arrow shows the time propagation done by applying the operator $P$ to $|x(t_i) \rangle$ which yields $|x(t_i + \Delta t) \rangle$. The blue arrows show the path where the data $|x(t_i) \rangle$ gets encoded into latent space, $|z_i \rangle$, then the operator $O$ is applied to the latent vector yielding a new latent vector $|z_j \rangle$. Finally, the new latent vector get decoded and yields $|x(t_i + \Delta t) \rangle$.} \label{fig:Prop}
\end{figure}

In the next section we introduce our mathematical methods and notation. In section \ref{sec:Model} we apply the method to the MNIST data set. In section \ref{sec:Class} we show how we can use this method  to understand the topology of the latent space. In section \ref{sec:Denoise} we apply this method to denoise images. In section \ref{sec:Op} we show how we can perform matrix operations in latent space which map to image transformations in real space.

\section{Method}
Assume a vector space which we call real space and denote the vectors in this space  $|x \rangle$ with $|x\rangle \in \Re^d$. Assume a set  $\lbrace |x_i\rangle \rbrace_{i=1}^N$, which we call the dataset with $N$ the dataset size. Similarly, we assume a vector space, which we call the latent space and denote these vectors  $|z\rangle$ with $|z\rangle \in \Re^M$ (in general, $M \leq d$). We also consider three deep neural networks, a Generator $\mathcal{G}$, an Encoder $\mathcal{E}$ and a Classifier $\mathcal{C}$. We can  interpret $\mathcal{G}$ as a projector from latent space to real space, \textit{i.e.}, $|x_i \rangle= \mathcal{G}| z_i \rangle$, and interpret $\mathcal{E}$ as the inverse of $\mathcal{G}$. However, given the architecture of variational autoencoders, notice that if $|z_a \rangle = \mathcal{E}|x_i \rangle$ and $|z_{a'} \rangle = \mathcal{E}|x_i \rangle$, in general, $|z_{a} \rangle \neq |z_{a'}\rangle$, since these vectors are  i.i.d. vectors sampled from a Gaussian distribution with mean and standard deviation dependent on $|x_i \rangle$ \cite{kingma2017variational}. Finally, the Classifier projects real-space vectors into the label space, \textit{i.e.}, $|y_k \rangle =\mathcal{C}|x_i \rangle$, where $|y_k \rangle \in L$, where $L$ denotes the label space. We assume that each vector $| y_k \rangle$ is a one-hot-vector. The length of $|y_k \rangle$ equals the number of labels $|L|=l$ and $k=1,...,l$. Henceforth, we assume that $l<M$.

We define  $\lbrace | \xi_i \rangle \rbrace_{i=1}^M$ to be a set of basis vectors in latent space such that $ \langle \xi_i | \xi_j \rangle = C \delta_{ij}$. Henceforth we call the set of basis vectors $\lbrace | \xi_i \rangle \rbrace_{i=1}^M$ the \textit{quasi-eigenvectors} since they form a basis and each one represents a feature \textit{state} in latent space. Notice that we can define the operator $\mathcal{A} = \sum_{j=1}^M | \xi_j \rangle \langle \xi_j |$, which implies $\mathcal{A} | \xi_i \rangle = C | \xi_i \rangle$.  Any vector in latent space can be expressed as a linear superposition of these quasi-eigenvectors, \textit{viz}, 
\begin{equation}
|z \rangle  = \sum_{j=1}^M c_j | \xi_j \rangle \; .
\label{eq:LatentSpaceDecomposition}
\end{equation}
where $|c_i| = |\langle \xi_i | z \rangle | $ is the amplitude of $| z \rangle$ with respect to $|\xi_i \rangle$ and gives a measure of $|z\rangle$'s projection with the quasi-eigenvector $|\xi_i \rangle$. Constructing a set of basis vectors is straightforward. However, we wish  each labeled feature to corresponds one-to-one with a quasi-eigenvector. Since we are assuming that $l<M$, there will be a set of quasi-eigenvectors that do not correspond to any labeled feature.


To obtain a set of orthogonal quasi-eigenvectors, we use the Gram-Schmidt method. Specifically:
\begin{enumerate}
    \item We train the GAN, using the training set $\lbrace |x_i \rangle \rbrace_{i=1}^N$ as in Ref. \cite{goodfellow2016nips}. \label{1}
    \item We train the Classifier independently, using the training set. \label{2}
    \item We train a VAE using the trained Generator as the decoder. 
    We also use the Classifier to classify the output of the VAE. We include in the loss function a regularizer $\lambda \cdot \mathcal{L}_{class}$, where $\lambda$ is a hyperparameter and $\mathcal{L}_{class}$ denotes the Classifier's loss function. At this stage, we only train the Encoder, keeping the Generator and Classifier  fixed.
    
    \item Define $n$ to be an integer such that $M = n \times l$. Then, for each label, we allocate $n$ sets of latent vectors and we denote them $|z_{\alpha,i}^k \rangle$, where $\alpha$ denotes the label, $i=1,...,n$ and $k=1,...,V$. Here $V$ is the number of elements (latent vectors) in each set corresponding to the pair $(i,\alpha)$. We build these sets $\lbrace |z_{\alpha,i}^k \rangle \rbrace$ in two ways: \label{5}
    \begin{enumerate}
        \item Using the training set, we encode each vector $|x_i \rangle \rightarrow |z_i \rangle = \mathcal{E} | x_i \rangle $, then we decode the latent vector, \textit{i.e.}, $|z_i \rangle \rightarrow |x_i \rangle = \mathcal{G} | z_i \rangle $, and then we classify the output, \textit{i.e.}, $|x_i \rangle \rightarrow |y_i \rangle = \mathcal{C}|x_i \rangle$. For each label $l$, there is a set of latent vectors. \label{3}
        \item We generate random latent vectors and map them to their labels using the Generator and the Classifier as in 4(a). \label{4} 
    \end{enumerate}
    We denote as $V$ the number of latent vectors in each set.
    \item We take the average over $V$ for each set of latent vectors $\lbrace |z_{\alpha, i} \rangle \rbrace_{k=1}^V$ and denote that average  $|\eta \rangle_{\alpha, i}$, \textit{i.e.}, \label{6}
    \begin{equation}
        |\eta \rangle_{\alpha, i} = \frac{1}{V}\sum_{j=1}^V |z_{\alpha, i}^j \rangle \; .
    \end{equation}
    Since the latent vectors are sampled from a multivariate Gaussian distribution, the average $|\eta_{\alpha, i} \rangle$ is finite and unbiased. By defining operators in latent space in terms of outer products of the $| \eta_{\alpha, i} \rangle$ vectors, these latent space operators will have encoded in them the set of latent vectors $|z_{\alpha,i}^k \rangle$. 
    \item To impose orthogonality, we use the Gram-Schmidt method. Thus, from the vectors $|\eta_{\alpha, i} \rangle$ we generate a set of quasi-eigenvectors $|\xi \rangle_{\alpha, i}$, \textit{i.e.}, \label{7}
    \begin{eqnarray}
        |\xi\rangle_{1, 1} &=& |\eta\rangle_{1, 1} \\
        |\xi\rangle_{2, 1} &=& |\eta\rangle_{2, 1} - \frac{_{2,1}\langle \eta |\xi\rangle_{1, 1}  }{_{1,1}\langle \xi |\xi\rangle_{1, 1}} |\xi\rangle_{1, 1} \\
        ... \\
        |\xi\rangle_{l, n} &=& |\eta \rangle_{l, n} - \sum_{\alpha=1}^{l-1}\sum_{i=1}^{n-1} \frac{_{l,n}\langle \eta |\xi\rangle_{\alpha, i}  }{_{\alpha, i}\langle \xi |\xi\rangle_{\alpha, i}} |\xi\rangle_{\alpha, i} \; .
    \end{eqnarray}
    Such that:
    \begin{equation}
        _{\alpha,i}\langle \xi | \xi \rangle_{\beta,j} = C \delta_{\alpha \beta} \delta_{ij} \label{eq:delta}
    \end{equation}
\end{enumerate}

In Eq. \eqref{eq:delta}, $C$ is the value of the norm. The set of quasi-eigenvectors $\lbrace | \xi \rangle_{\alpha,i} \rbrace_{\alpha=1, i=1}^{l,n}$ span the latent space and, as we will show, a subset of them map to specific features. 

The key point is that the set of quasi-eigenvectors form a basis set in latent space and each direction corresponds to a feature in real space. This structure allows us to give a better topological description of latent space, \textit{i.e.}, how does labeled features map to latent space similar to how molecular configurations map to the \textit{energy landscape} \cite{wales2003energy}. In addition, we can use the set of quasi-eigenvectors as tools for classification, denoising and topological transformations. We  demonstrate these applications in the next section using the MNIST dataset.

\section{Model} \label{sec:Model}

\begin{figure}[hbtp]
\centering
\includegraphics[width=3.2in, height=1.in]{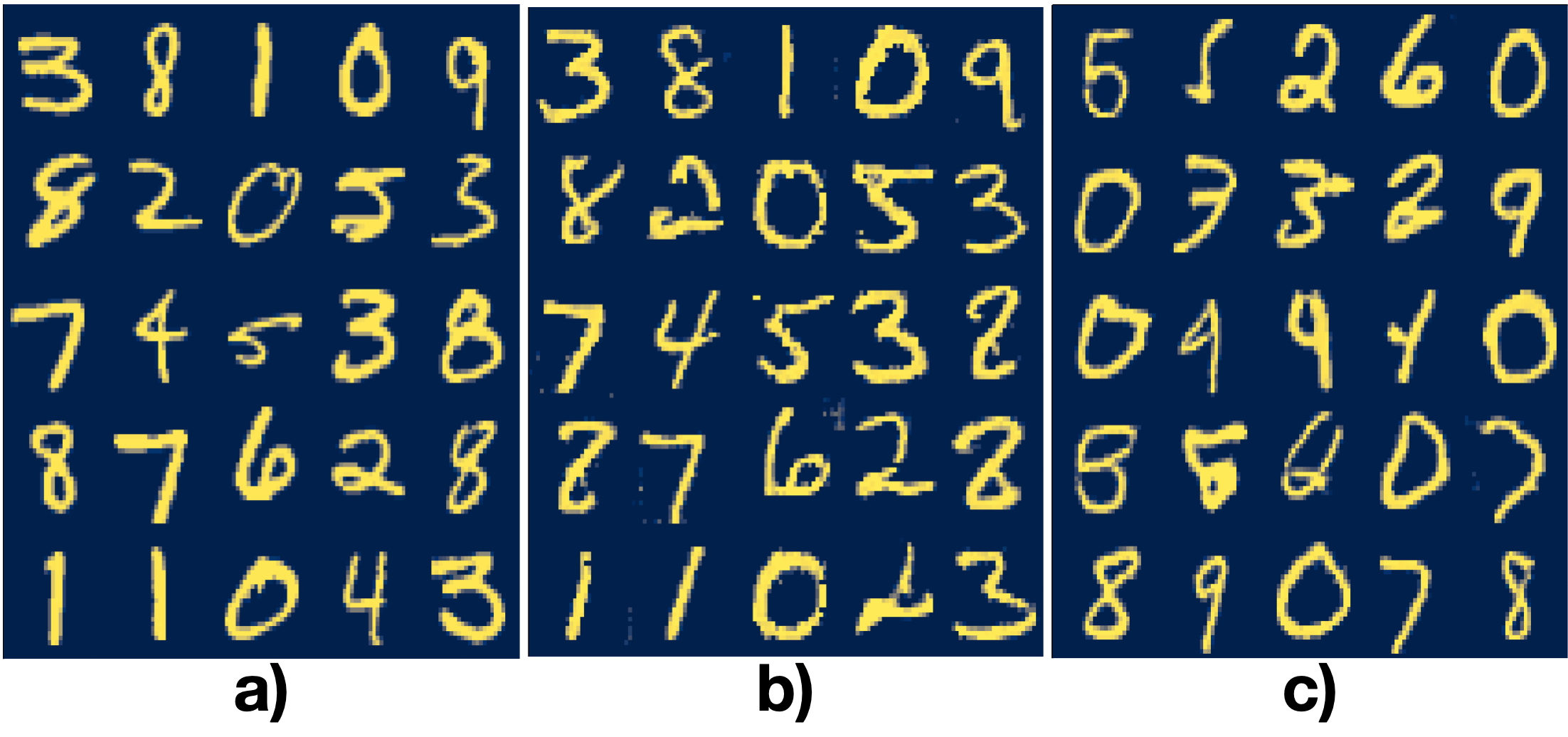}
\caption{ A batch from the \textbf{a}) dataset, \textbf{b}) the same dataset encoded and decoded using the Generator as Decoder and \textbf{c}) random latent vectors given as input to the Generator.} \label{fig:NNs}
\end{figure}

 We trained a GAN, a Classifier and a VAE using the MNIST dataset which has $60k$ and $10k$ one-channel images in the training and test set, respectively, with dimensions $28\times28$ pixels. We fixed the batch size to 25 and number of epochs to 500 during all training runs. We trained the GAN using the training set,  used the Jensen-Shannon entropy as the loss function \cite{goodfellow2016nips}, the ADAM optimizer with hyperparameters $\eta=0.0002,\; \beta_1 = 0.9, \; \beta_2=0.999$ for both the Generator and the Discriminator, fixed the latent space dimensionality to $M=100$ and sampled the random latent vectors from a multivariate Gaussian distribution centered at the origin  with standard deviation equal to $1$ in all $M$ dimensions. Independently, we trained a Classifier on the training set, used  crossentropy as loss function and a softmax as the activation function in the last layer, the ADAM optimizer with hyperparameters $\eta=3\cdot 10^{-5},\; \beta_1 = 0.5, \; \beta_2=0.99$. The accuracy of the classifier on the test set reached $\approx 98.9\%$. Using the training set, we then trained the Encoder in a VAE and  used the Generator as the Decoder. We used as loss function the Kullback–Leibler divergence and the hinge loss function. We also added as a regularizer the Classifier's loss function and the Lagrange multiplier, $\lambda$, as hyperparameter set to $\lambda=100$. During the training of the Encoder, we kept both the Generator and the Classifier  fixed.  In Fig. \ref{fig:NNs} we show the training results. To train the NNs we used Flux \cite{innes2018flux} in Julia \cite{Julia} and the code can be found in Ref. \cite{githubJaque}.
 
\begin{figure}[hbtp]
\centering
\subfigure[]{
\includegraphics[width=3.0in, height=2.2in]{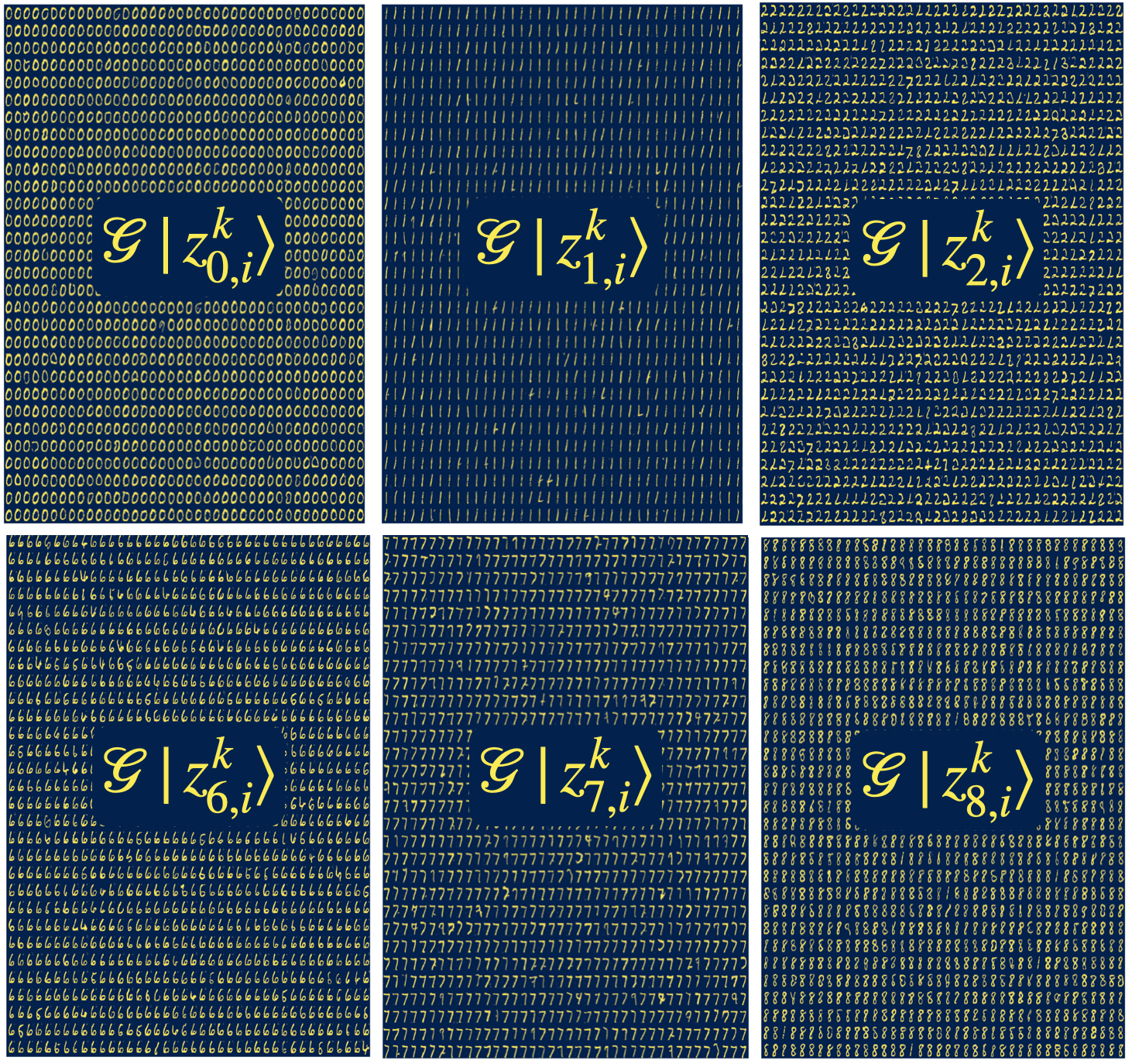}
\label{fig:Samples012678}}
\centering
\subfigure[]{
\includegraphics[width=3.0in, height=1.4in]{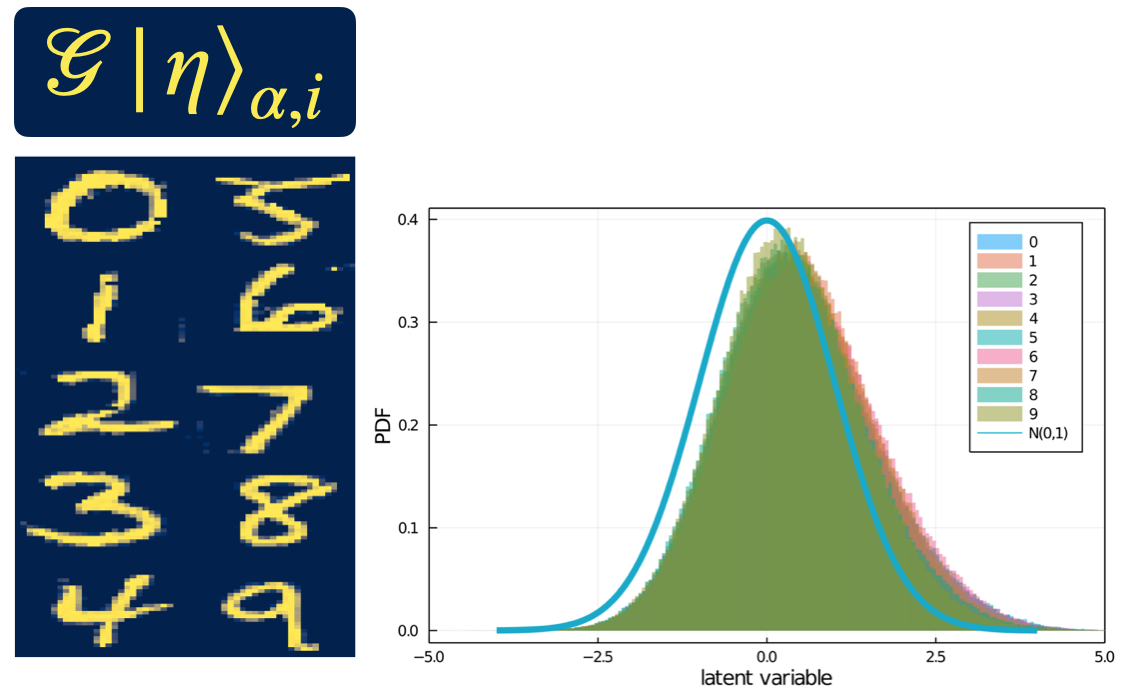}}
\label{fig:SupNumEnc}
\centering
\subfigure[]{
\includegraphics[width=3.0in, height=1.4in]{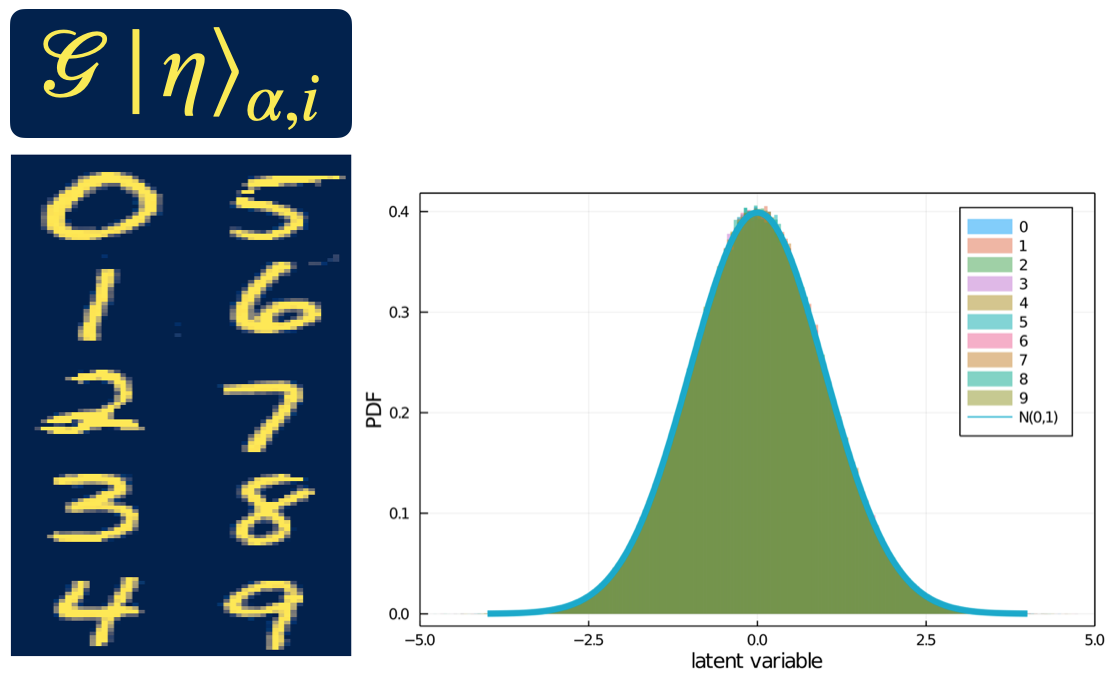}}
\label{fig:SupNum}
\caption{\textbf{a}) Samples of latent vectors $|z_{\alpha, i}^k \rangle$, for labels $\alpha=0,1,2,6,7$ and $8$. $\mathcal{G}|z_{\alpha, i}^k \rangle$ yields images of numbers with label $\alpha$. We show 1200 latent vectors, per label, projected into real space. Average over the latent vectors per label yields $|\eta\rangle_{\alpha, i}$. \textbf{b) left panel} Decoded latent vectors $|\eta\rangle_{\alpha, i}$. The vectors $|\eta\rangle_{\alpha, i}$ were obtained as described in step 4(a). \textbf{b) right panel} The histogram for each label is Gaussian with non-zero mean. \textbf{c) left panel} Decoded latent vectors $|\eta\rangle_{\alpha, i}$. The vectors $|\eta\rangle_{\alpha, i}$ were obtained as described in step 4(b). \textbf{c) right panel} The histogram for each label is Gaussian with zero mean.} \label{fig:Hist}
\end{figure}

The latent space dimension is $M=100$, while the number of labels is $|L|=10$. Thus, following step 4, for each label we generated $n=M/|L|$ sets of latent vectors, each set containing $V=5000$ latent vectors. In Fig. \ref{fig:Samples012678} we show sample  latent vectors for labels $0,1,2,6,7$ and $8$, projected to real space using the Generator $\mathcal{G}$. Then we take the average over each set as in step 5. We checked that the average and standard deviation over each of the entries in the set of vectors $\lbrace |\eta \rangle_{\alpha, i} \rbrace_{\alpha,i}$ converges.  Interestingly, when taking the average over the set of latent vectors corresponding to a label and projecting back to real space, the label holds. For instance, in Fig. \ref{fig:SupNumEnc} we show the projected image of the average over $V$ for each set of latent vectors $\lbrace |z_{\alpha, i} \rangle \rbrace_{k=1}^V$ in the case where the latent vectors were obtained following step 4(a), whereas Fig. \ref{fig:SupNum} corresponds to the case following step 4(b). We have also plotted the probability density function (\textit{PDF}) per label in latent space for both cases and added a Gaussian distribution with mean and standard deviation equal to $0$ and $1$, respectively, for reference. Notice that the PDF in Fig. \ref{fig:SupNumEnc} is  shifted away from the Normal distribution, whereas in Fig. \ref{fig:SupNum} all PDFs are bounded by the Normal distribution, because latent vectors generated directly from latent space are, by definition, sampled from a multivariate Gaussian distribution with mean and standard deviation equal to $0$ and $1$, respectively. On the contrary, encoding real space vectors yields Gaussian vectors overall (\textit{i.e.}, the PDF over all latent vectors over all labels yields a Gaussian distribution, by definition) but the mean and standard deviation can differ from $0$ and $1$ \cite{kingma2017variational}. 

Step 4(a) gives robustness to this method and step 4(b) allows us to generate as many latent vectors as we want with a specific label. Since the latent space dimension is $M=100$, we need $M$ averaged latent vectors $|\eta\rangle_{\alpha, i}$ to generate $M$ orthogonal latent vectors. Since the number of labels is $\alpha=\lbrace 0,...,|L|-1 \rbrace$, then $n=10$. To this end, we generate one set (\textit{i.e.}, $i=1$) following step 4(a) and nine sets (\textit{i.e.}, $i=2,3,...,n$) following step 4(b).

\begin{figure}[hbtp]
\centering
\subfigure[]{
\includegraphics[width=1.5in]{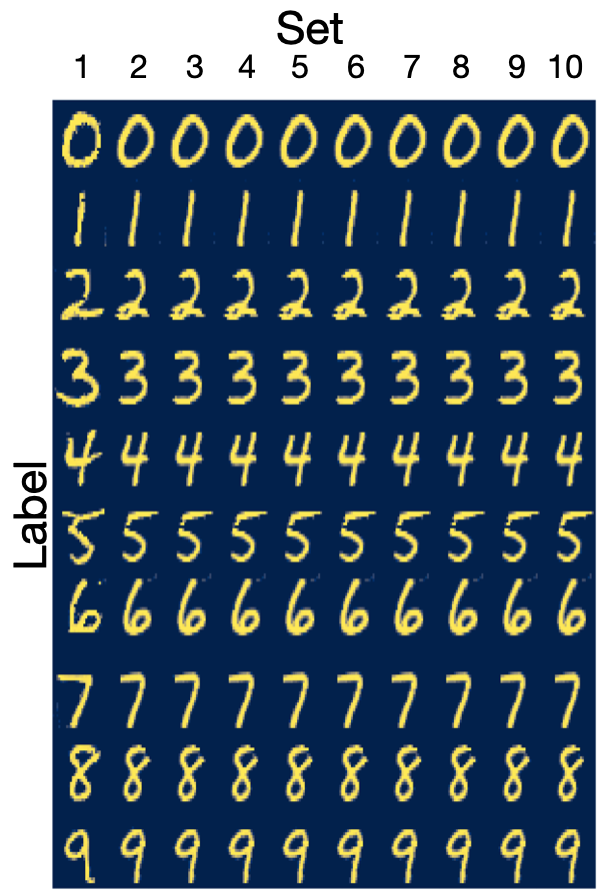}
\label{fig:Vs-numbers}}
\centering
\subfigure[]{
\includegraphics[width=1.5in]{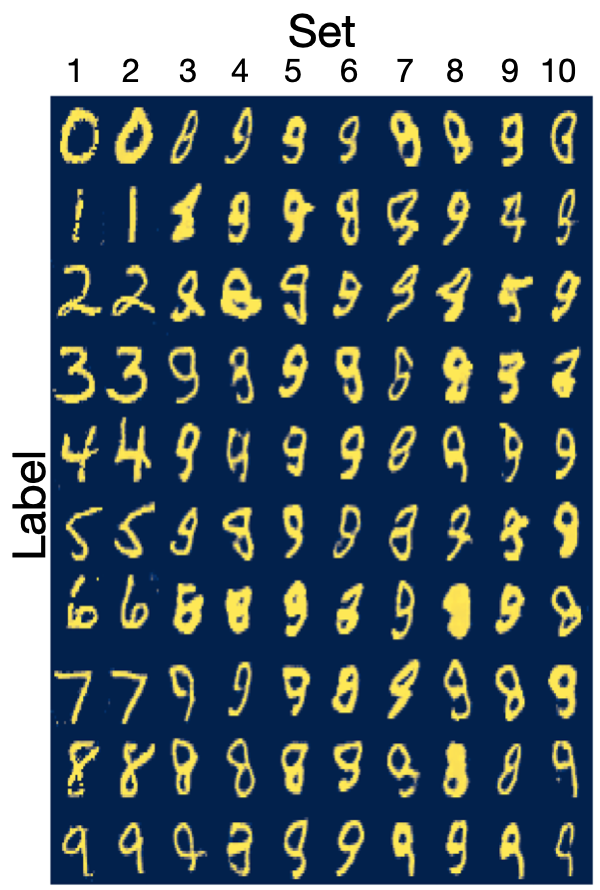}
\label{fig:Us-numbers}}
\centering
\subfigure[]{
\includegraphics[width=1.5in,height=1.2in]{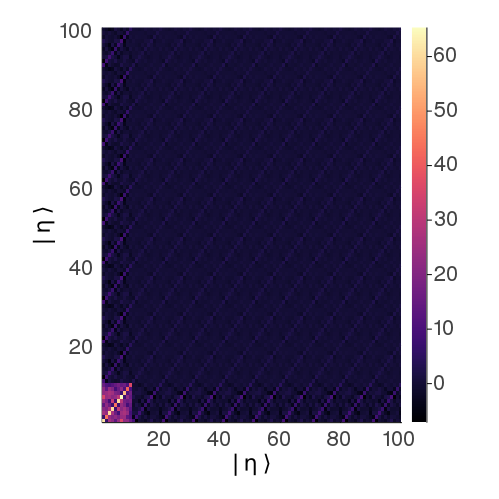}
\label{fig:Vs-heatmap}}
\centering
\subfigure[]{
\includegraphics[width=1.5in, height=1.2in]{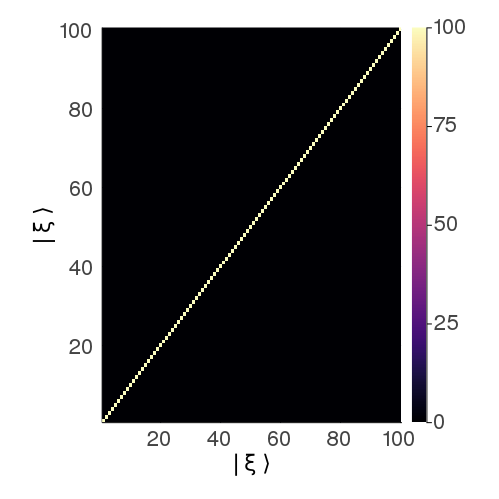}
\label{fig:Us-heatmap}}
\caption{ \textbf{a)} Projection to real space images of the latent vectors $\lbrace |\eta \rangle_{\alpha, i} \rbrace_{\alpha=0, i=1}^{9,10}$ obtained as described in step 5. \textbf{b)} Projection to real space images of the quasi-eigenvectors $\lbrace |\xi \rangle_{\alpha, i} \rbrace_{\alpha=0, i=1}^{9,10}$ obtained as described in step 6. The $\alpha$ index corresponds to the label (row) while the $i$ index correspond to the set (column). \textbf{c)} The inner product of vectors $\lbrace |\eta \rangle_{\alpha, i} \rbrace_{\alpha=0, i=1}^{9,10}$. \textbf{d)} The inner product of the quasi-eigenvectors $\lbrace |\xi \rangle_{\alpha, i} \rbrace_{\alpha=0, i=1}^{9,10}$.} \label{fig:VsandUs}
\end{figure}

Fig. \ref{fig:Vs-numbers} shows the projection to real space of all the $|\eta \rangle_{\alpha, i}$ vectors while Fig. \ref{fig:Vs-heatmap}  shows the inner product $_{\alpha, i}\langle \eta |\eta \rangle_{\alpha', i'}$ as a heatmap, which shows they are non-orthogonal. 
At this point, we have $M$ vectors $| \eta \rangle_{\alpha, i}$ in latent space i) composed of the sum of $V$ latent vectors, ii) each  of these vectors maps to a specific feature. However, these vectors are not orthogonal.
Using the Gram-Schmidt method described in step 6, we obtain a set of vectors, $|\xi \rangle_{\alpha, i}$, in latent space such that i) each $|\xi \rangle_{\alpha, i}$ vector encodes $V$ latent vectors, ii) each $|\xi \rangle_{\alpha, i}$ vector maps to a specific labeled feature (see Fig. \ref{fig:Us-numbers}), iii) the $|\xi \rangle_{\alpha, i}$ vectors are orthogonal, as shown in Fig. \ref{fig:Us-heatmap}. Since the Generator was trained using random vectors sampled from a multivariate Gaussian distribution centered at zero  with standard deviation  $1$,  the value of the norm of any random latent vector will be $\langle z | z \rangle \approx M$. Therefore, we fixed the norm of the quasi-eigenvectors to be $C = M$ (see Eq. \eqref{eq:delta}).

Notice that while the non-orthogonal vectors $| \eta \rangle_{\alpha, i}$ for the MNIST GAN project sharp images of easily-identifiable numbers in real space, not all quasi-eigenvectors project to images of numbers in real space. Only a few of the $M$ linearly-independent directions in latent space ($\approx 20$) project to images of numbers in real space. We will show how to apply this property of the quasi-eigenvectors to the MNIST test set to classify images in latent space and to denoise real-space images. We also show how to build a rotation operator in latent space that generates feature transformations in real space. 


\section{Using LSD as a classifier in latent space} \label{sec:Class}
We can express any latent vector $|z\rangle$,  in terms of the quasi-eigenvectors, \textit{viz.}
\begin{equation}
    |z\rangle = \sum_{k=1}^M c_k | \xi_{k} \rangle \; ,
\end{equation}
where the coefficients $c_k$ are given by,
\begin{equation}
    c_k = \langle \xi_k | z \rangle / C \; .
\end{equation}

\begin{figure}[hbtp]
\centering
\subfigure[]{
\includegraphics[width=1.5in]{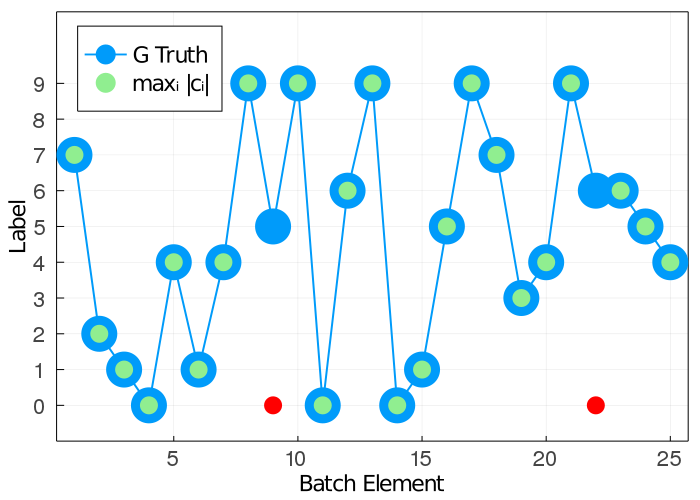}
\label{fig:true_labels_a}}
\subfigure[]{
\includegraphics[width=1.5in]{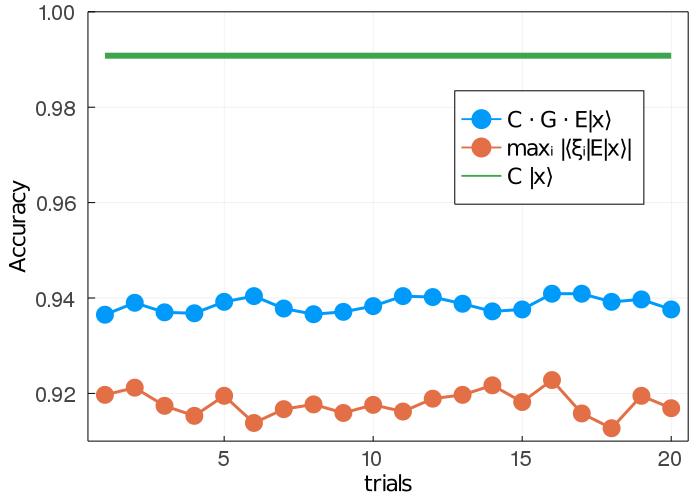}
\label{fig:Accuracy_comparison}}
\centering
\subfigure[]{
\includegraphics[width=3.1in]{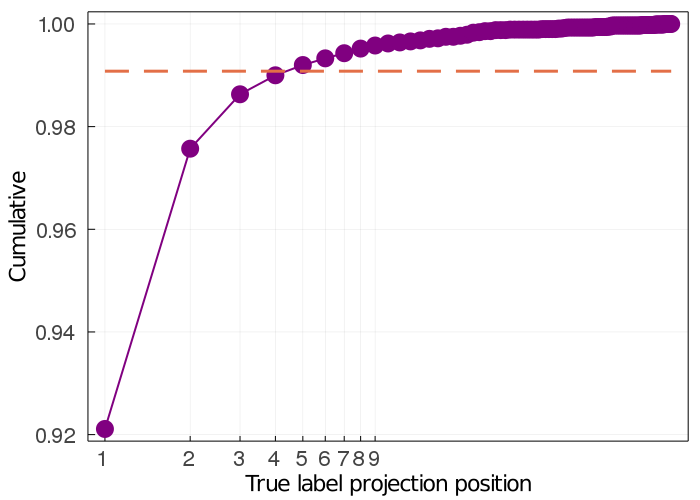}
\label{fig:cumulative}}
\caption{ \textbf{a)} A batch of the MNIST test set classified by LSD using the largest amplitude. The largest amplitude $|c_i|$ corresponds to the quasi-eigenvector $|\xi_i \rangle$ that contributes the most to the latent vector $|z\rangle$, and a subset of the quasi-eigenvectors map to each label one-on-one. Y axis corresponds to the label, X axis to the image in the batch. Blue dots, ground truth. Green (red) dots correspond to the case(s) where the label associated with the quasi-eigenvector with the highest amplitude is the correct (incorrect) label. \textbf{b)} Accuracy for different trials using the MNIST test set. The green curve is the Classifier's accuracy ($98.9\%$),  the blue dots are the accuracy over the encoded-decoded MNIST test set ($\approx 94\%$) and the red dots corresponds to the accuracy using the the largest amplitude in LSD ($\approx 92\%$). \textbf{c)} Cumulative probability of the ground truth label being any of the $n$ first largest amplitudes (X axis). For $n=1$ the probability is $92\%$. 
The probability of the ground truth label being one of the labels with the 4 largest amplitudes is $\approx 98.9\%$, which is the classifiers accuracy.  } \label{fig:Amp}
\end{figure}

Similar to principal component analysis, we are interested in how much information about an image is encoded in the quasi-eigenvector with the largest amplitude $|c_i|$. We encode images from the MNIST test set into latent space, then express the latent vectors in terms of the quasi-eigenvectors (we call this expression \textit{latent spectral decomposition} or \textit{LSD}) and find the maximum amplitude $|c_i|$ for each latent vector. Recall that the amplitude $|c_i|$ is a measure of the projection of the latent vector with respect to the quasi-eigenvector $|\xi_i \rangle$. Thus, the largest amplitude corresponds to the quasi-eigenvector that contributes the most to the latent vector. Since the quasi-eigenvectors are associated with labeled features in real space, we use the largest amplitude as a way to classify the image. Fig. \ref{fig:true_labels_a} shows a sample batch of 25 images. The blue dots corresponds to the true labels (see y axis), while the green (red) dots correspond to the case where label associated with the quasi-eigenvector with the largest amplitude is the correct (incorrect) label. In this batch, only batch elements $9$ and $22$ have true labels that do not agree with the label of the quasi-eigenvalue of the image with the largest amplitude. Since each time the Encoder encodes an image it generates a new random latent vector, then we could obtain a different outcome for batch elements $9$ and $22$ as well as the rest of the batch elements for each trial. For this reason, we perform an ensemble average over $20$ trials. For each trial we take the whole MNIST test set and compute the accuracy of the latent space decomposition (LSD) classifier (see red dots in Fig. \ref{fig:Accuracy_comparison}). We also computed the accuracy when the test set is encoded through the Encoder, then decoded through the Generator and finally classified (see blue dots in Fig. \ref{fig:Accuracy_comparison}). We have included the accuracy of the trained Classifier in Fig. \ref{fig:Accuracy_comparison} as an upper bound. While the trained Classifier has an accuracy of $98.8\%$, the LSD classifier has an average accuracy of $\sim 92\%$. This difference in accuracy, however, should not be interpreted as showing that the latent-space classifier does a poor job, but that the dominant few quasi-eigenvectors carry most of the information in latent space regarding the individual test-set images. In fact, the encoded $99\%$ of the test-set data requires only the 10 linearly-independent directions in set 1, i.e., the largest amplitude correspond to quasi-eigenvectors in the first set. 

\begin{figure}[hbtp]
\centering
\includegraphics[width=3.2in]{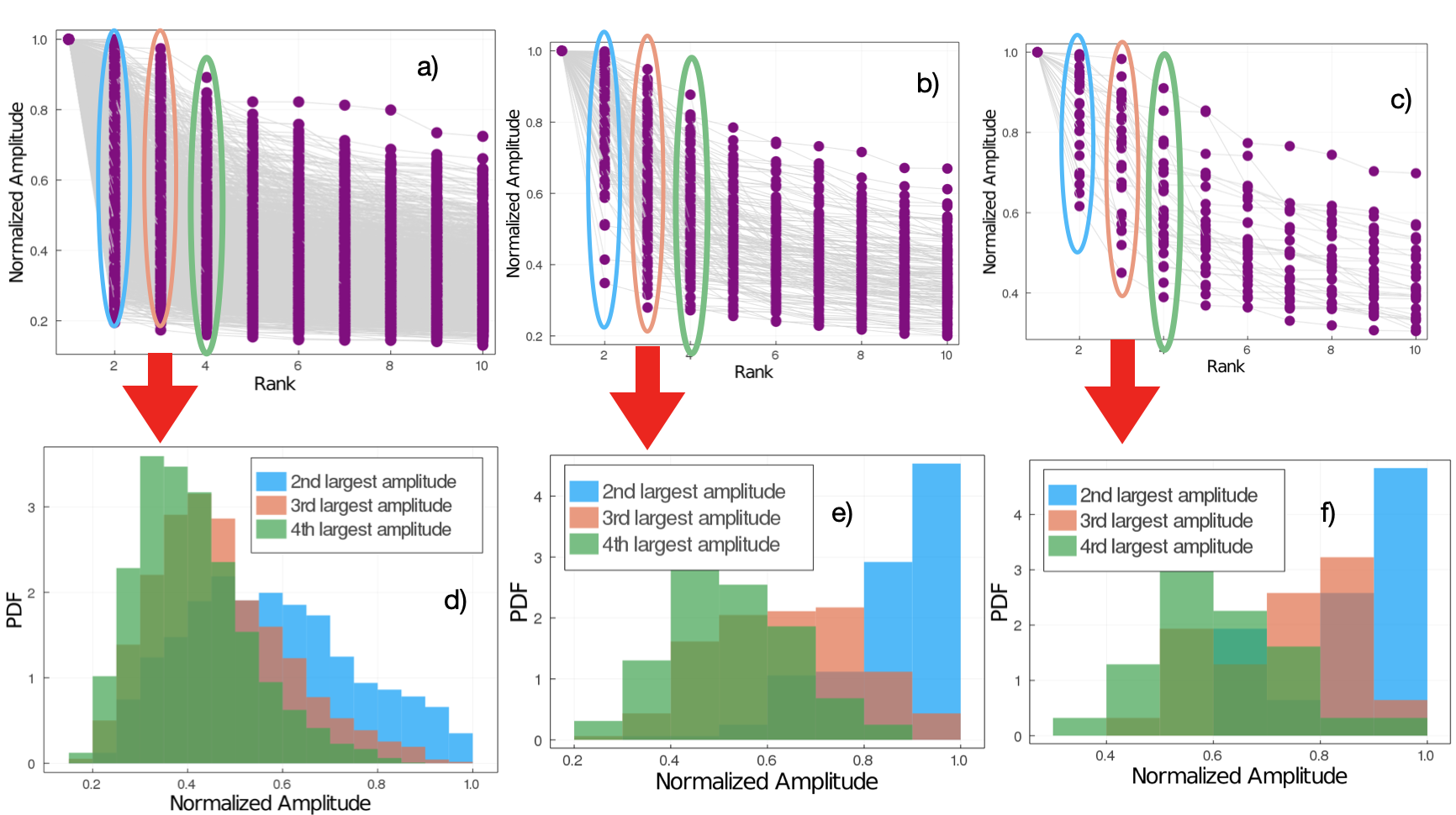}
\caption{ LSD normalized amplitude ranking for the cases where the true label corresponds to the largest amplitude quasi-eigenvector \textbf{a)}, second largest amplitude \textbf{b)} and third largest amplitude \textbf{c)}. Probability-density function of the second-, third- and fourth-largest LSD normalized amplitude when the true label corresponds to the largest amplitude \textbf{d)}, second-largest amplitude \textbf{e)} and third-largest amplitude \textbf{f)}.} \label{fig:Rank}
\end{figure}

Suppose that when we perform the LSD, we sort the amplitudes such that $|c_1| > |c_2| > ... > |c_M|$ and ask \textit{the position of the ground-truth label?} As previously mentioned, in $92\%$ of the cases the ground-truth label corresponds to the first position (\textit{i.e.}, $|c_1|$). In $5\%$ of the cases the ground truth label corresponds to the second largest amplitude (\textit{i.e.}, $|c_2|$). In Fig. \ref{fig:cumulative} we have plotted the cumulative of the probability for the ground-truth label being any of the first $n$  positions. The dashed red line corresponds to the trained Classifier accuracy. Notice that the probability of the label being in position 1, 2, 3 or 4 of the LSD equals the accuracy of the trained classifier, \textit{i.e.}, in $98.9\%$ of the MNIST test-set images the ground truth label is associated to a quasi-eigenvector such that the associated coefficient is either $c_1, c_2, c_3$ or $c_4$. In this sense, it is possible that even when the amplitude of the quasi-eigenvector associated to the ground-truth label is not the largest one, rather the 2nd or 3rd largest one, then $|c_1| \gtrsim |c_2|$ or $|c_1| \gtrsim |c_2| \gtrsim |c_3|$. To test this idea, in Fig. \ref{fig:Rank} we have plotted the normalized amplitude (\textit{i.e.}, $|c_i|/\max \lbrace |c_j| \rbrace$) \textit{vs} the rank (\textit{i.e.}, sorted amplitudes from largest to smallest) for all images in the test set. Fig. \ref{fig:Rank} a) corresponds to the images where the LSD amplitude of the quasi-eigenvector associated with the ground-truth label is the largest, whereas in Fig. \ref{fig:Rank} b) and c) the amplitude is the 2nd largest or 3rd largest, respectively. Given the large dataset, in Fig. \ref{fig:Rank} d), e) and f) we have plotted the PDFs of the 2nd, 3rd, and 4th largest amplitudes for each of  plots Fig. \ref{fig:Rank} a), b) and c). To be clear, from Fig. \ref{fig:Rank} a), b) and c) we generated PDFs for the second-, third- and fourth-largest amplitudes in each plot and  show the PDFs in Figs. \ref{fig:Rank} d), e) and f), respectively. Notice that when the largest amplitude corresponds to the ground-truth label (Fig. \ref{fig:Rank} a)), the second-, third- and fourth-largest amplitude PDFs are centered below 0.6 (Fig. \ref{fig:Rank} d)). When the second-largest amplitude corresponds to the ground-truth label (Fig. \ref{fig:Rank} b)) the PDF of the second-largest amplitude is shifted towards 1, while the PDFs of the third- and fourth-largest amplitude amplitudes are centered below 0.7 (Fig. \ref{fig:Rank} e)). Finally, in the case where the third-largest amplitude corresponds to the ground-truth label (Fig. \ref{fig:Rank} c)), the PDFs of the second- and third-largest amplitude are shifted towards 1, while the PDF of the fourth-largest amplitude is centered below 0.7 (Fig. \ref{fig:Rank} f)).

The previous results give us a broad picture of  latent space topology: the labeled features project to well-defined compact domains in latent space. Let us now consider how we can use this information to denoise images.

\section{Denoising with LSD} \label{sec:Denoise}
The main issue when reducing noise in images is distinguishing  noise from information. In this sense, a reliable denoiser has to learn what is noise and what isn't. 
One reason deep generative models are promising for denoising data is that in optimum conditions the GM has learned the exact data distribution. Of course, if the data set has noise, the GM will also learn the embedded noise in the data set. However, by sampling the latent space we may find regions where the signal to noise ratio is sufficiently large. For large $M$, this sampling is computationally expensive. To avoid this cost, we propose to LSD as a denoiser.

\begin{figure}[hbtp]
\centering
\subfigure[]{
\includegraphics[width=2.in]{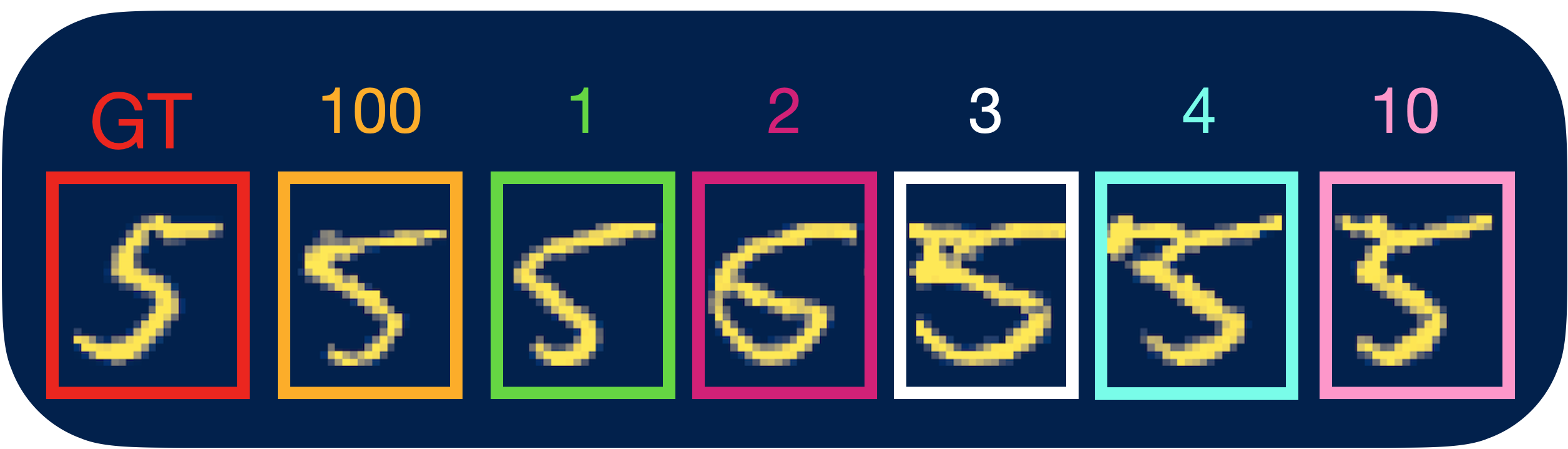}
\label{fig:denoise-samp}}
\centering
\subfigure[]{
\includegraphics[width=3.2in, height=2.0in]{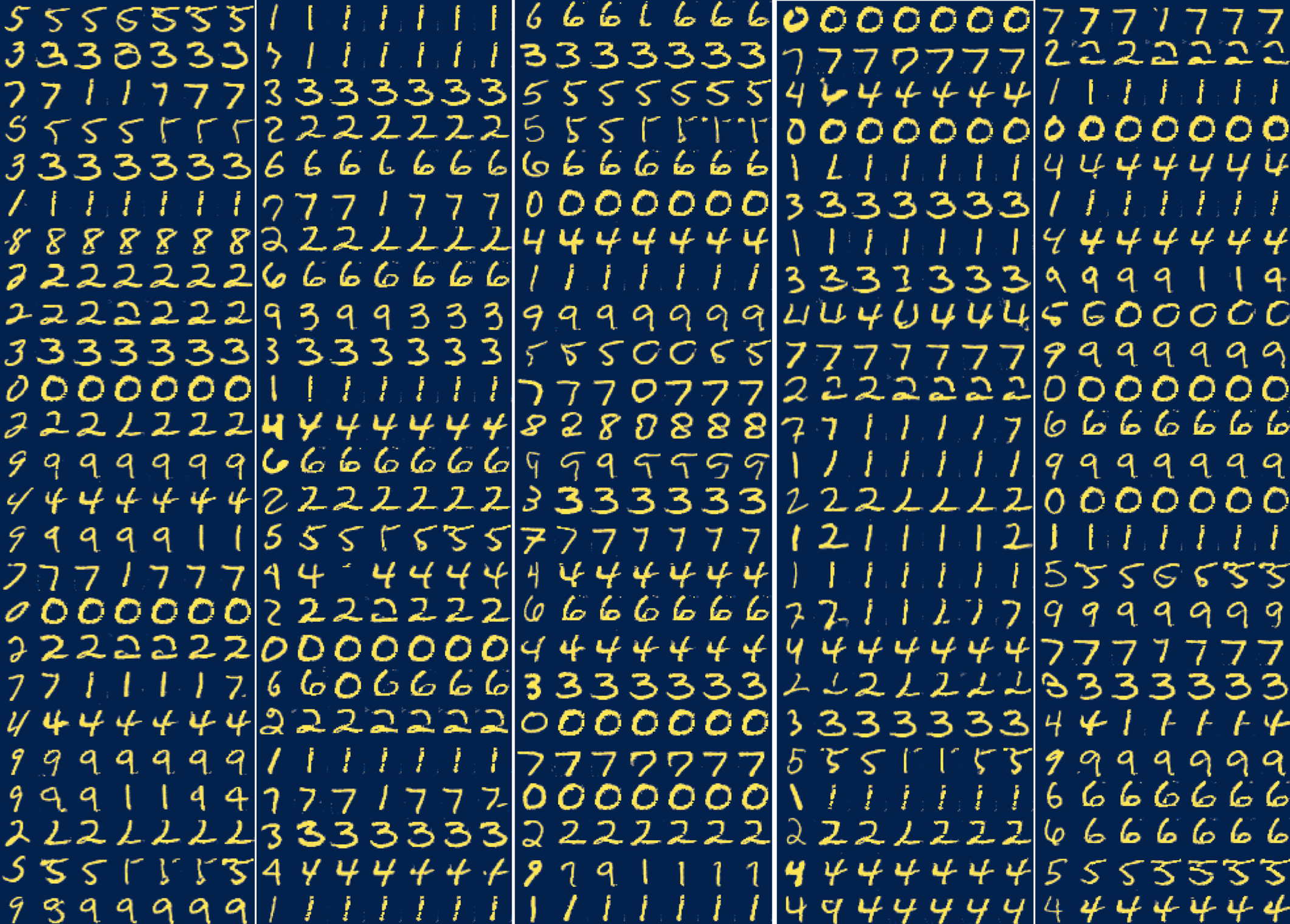}
\label{fig:denoiser2}}
\caption{\textbf{a)} Image of the number 5 taken from the MNIST test set. The first image correspond to the ground truth (\textit{GT}), the second image corresponds to the projected image of the 100 LSD components, the third, fourth, fifth, sixth and seventh images correspond to the projected images from the sum of the one, two, three, four and ten LSD components, respectively. \textbf{b)} 125 samples from the MNIST test set. Each sample is a row with seven images as shown in \textbf{a)}.} \label{fig:denoiser}
\end{figure}

Recall that in the previous section we showed that with a $98\%$ accuracy the information needed to assign a label to the image is stored in either the first-, second-, third- or fourth-largest amplitude of the LSD. Therefore, we propose that once the test set is encoded into latent space, we decompose the latent vector in terms of the quasi-eigenvectors and drop the contribution from quasi-eigenvectors with low amplitudes. In Fig. \ref{fig:denoiser} we show the results of this truncation for 125 random sample images. In Fig. \ref{fig:denoise-samp} we describe how to understand these images. Fig. \ref{fig:denoiser2}  shows 5 columns, where each column has 25 rows and each row has 7 images. In each row, the first image corresponds to the ground-truth image, the second image is the  image decoded from all 100 LSD components of the ground truth image. The third, fourth, fifth, sixth and seventh images are the  images decoded after truncating the expansion after 1,2,3,4 and 10 LSD components of the ground truth image.
In this method, denoising maintains the identity of the labeled feature in the image, \textit{e.g.}, each row shows different representations of the same number. In most cases in Fig. \ref{fig:denoiser}, the denoised image looks clearer and sharper. However, sometimes the LSD components project back to the wrong number. However we can consider as many LSD components as the dimension of the latent space, so even if taking the first $n$ LSD components yields the wrong number,  taking the first $n+1$ LSD components could yield the correct number. In the previous section we showed that using only the first $4$ LSD components gave us a $98.9\%$ chance of obtaining the right number.


\section{Operations in latent space} \label{sec:Op}
Here we explore how to build operators in latent space that can yield feature transformations in real space. Having a set of orthogonal vectors that span latent space allows us to perform most operations in latent space as a series of rotations, since we can express the operator as a superposition of the outer product of the quasi-eigenvectors. If we construct a rotation matrix, $\mathcal{R}$, in latent space, we can then recursively apply $\mathcal{R}$ to a set of encoded images. After each iteration we project the output to real space to see the effect of the latent-space rotation. We can define a projection operator $\mathcal{B}_{\xi_i, \xi_j}$, such that, 
\begin{equation}
    B_{\xi_i,\xi_j} = \frac{1}{\langle \xi_i | \xi_i \rangle} | \xi_j \rangle \langle \xi_i | \; .
\end{equation}
This operator projects from $|\xi_i \rangle$ to $| \xi_j \rangle$, i.e., $\mathcal{B}_{\xi_i, \xi_j} | \xi_k \rangle = \delta_{\xi_k, \xi_k} | \xi_j \rangle$, where $\delta_{\xi_k, \xi_k}$ denotes the Kroenecker delta function.
Similarly, we define the operator $\mathcal{R}_{\xi_i, \xi_j}(\Delta \theta, \theta)$ as
\begin{multline}
    \mathcal{R}_{\xi_i, \xi_j}(\Delta \theta, \theta) \propto \left(\cos(\theta + \Delta \theta) | \xi_i \rangle + \sin(\theta + \Delta \theta) | \xi_j \rangle \right) \\ 
    \cdot \left(\langle \xi_i | \cos(\theta) + \langle \xi_j | \sin(\theta)  \right) \; ,
\end{multline}
which projects from $\cos(\theta ) | \xi_i \rangle + \sin(\theta ) | \xi_j \rangle$ to $\cos(\theta + \Delta \theta) | \xi_i \rangle + \sin(\theta + \Delta \theta) | \xi_j \rangle$.

\begin{figure}[hbtp]
\centering
\rotatebox[origin=c]{0}{\includegraphics[width=3.2in]{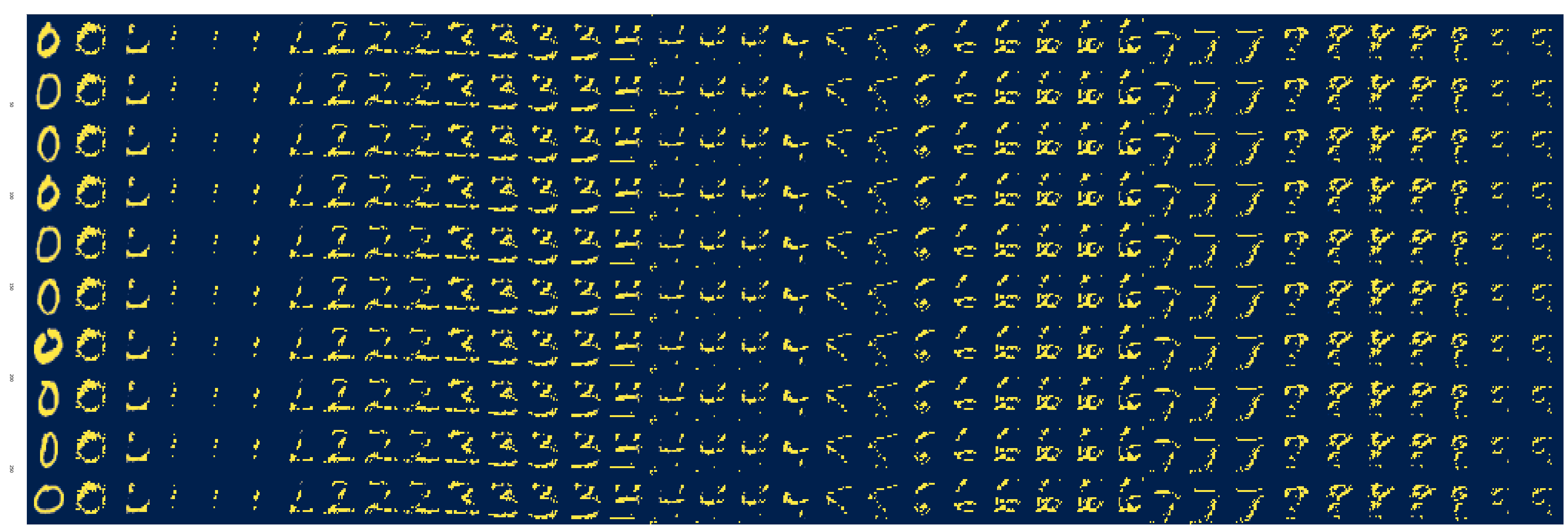}}
\caption{ Ten latent vectors projected into real space after each iteration where the latent vectors are rotated an angle $\Delta \theta \approx \pi/6$ from between two linearly independent directions associated with a quasi-eigenvector each. Rotations in latent space map to real space as label feature transformation, \textit{i.e.}, the images transform from the number $0$ to number $1$ and then from $1$ to $2$ until reaching number $9$.} \label{fig:rotations}
\end{figure}

Starting from a set of images with label zero, we first encoded them to latent space, then we applied the rotation operator $\mathcal{R}$ recursively, as follows: First, we perform the rotation from the quasi-eigenvector associated with label zero to the quasi-eigenvector associated with label 1, \textit{viz.}, $\mathcal{R}_{\xi_{\alpha=0, i=1}, \xi_{\alpha=0, i=1}}(\Delta \theta, \theta)$. Then, we performed a rotation from the quasi-eigenvector associated with label 1 to the quasi-eigenvector associated with label 2, \textit{viz.}, $\mathcal{R}_{\xi_{\alpha=1, i=1}, \xi_{\alpha=2, i=1}}(\Delta \theta, \theta)$, and repeat \textit{mutatis mutandi } until we reach the quasi-eigenvector associated with label $\alpha=9$. To keep the individual rotations in latent space small (and maintain the local linearity of the transforms), we fixed the rotation step size $\Delta \theta \approx \pi/6$ so transforming from a direction associated with one quasi-eigenvector to a direction associated with a different quasi-eigenvector requires three sequential rotations. In Alg. \ref{alg:0} we show the pseudocode. To ensure the rotated latent vectors have constant norm value as in Eq. \eqref{eq:delta}, after each iteration we divide the latent vector $|z\rangle$ by $\sqrt{\frac{\langle z | z \rangle}{M}}$. After each iteration, we project the latent vector into real space. In Fig. \ref{fig:rotations} we show this projection for a set of sample images. Notice how the numbers transform from $0$ to $9$. In principle, we could rotate through any other set of sequential features in this way. The key idea is that having a set of quasi-eigenvectors that span latent space each mapping to a specific label, we can define a metric in latent space defining the distance between the latent-space representation of each label.

\begin{algorithm}[hbtp]
\SetAlgoLined
 initialization \\
 $| z \rangle = \mathcal{E} | x \rangle $ \qquad (initial condition) \\
 $\Delta \theta = \pi/3$ \qquad (angular rotation step) \\ 
 $\alpha = 0$ \qquad\qquad (initial label) \\
 $i=1$ \qquad \qquad (set index) \\
 \For{$\alpha$ in $\lbrace 0,1,2,...,9 \rbrace$}{
  \For{$r$ in $\lbrace 1,2,3 \rbrace$}{
  $| z \rangle = \mathcal{R}_{\xi_{\alpha, i}, \xi_{\alpha,i}}(r \cdot \Delta \theta, (r-1) \cdot \Delta \theta) |z\rangle $ (rotation) \\ 
  $| z \rangle = | z \rangle /\sqrt{\frac{\langle z | z \rangle}{M}}$ \qquad (norm) \\
  $|x \rangle = \mathcal{G}|z\rangle$ \qquad (projection to real space)
  }
 }
 \caption{Latent space rotation pseudocode.} \label{alg:0}
\end{algorithm}


\section{Conclusions} \label{sec:Conclusions}
We have shown that it is possible to build a set of orthogonal vectors (quasi-eigenvectors) in latent space that both span latent space and map to specific labeled features. These orthogonal vectors reveal the latent space topology. 
We found that for MNIST, almost all the images in the data set  map to a small subset of the dimensions available in latent space.
We have shown that we can use these quasi-eigenvectors to reduce noise in data. We have also shown that we can perform matrix operations in latent space that map to feature transformations in real space. 

On the one hand, the deeper the NN the better its capacity in learning complex data and as depth increases, the non-linearity increases as well. From catastrophe theory \cite{arnold2013dynamical}, we know that in non-linear dynamical systems small perturbations can be amplified leading to bifurcation points leading to completely different solution families of these non-linear dynamical systems.
On the other hand, the results in Ref. \cite{radford2015unsupervised}  suggest a different picture with what the authors call \textit{vector arithmetics}. This arithmetic occurs in latent space. Similarly, there's an ongoing debate on how deep should a NN be to perform a specific task. In addition, it has been proposed the equivalence between deep NNs and shallow wide NNs \cite{bahri2020statistical}. 
 Our work contributes to this discussion of the emergent effective linearity of NNs as transformations. While the NNs we used are intrinsically non-linear, they exhibit local linearity over a region of interest in latent space. This subspace maps to labeled features. In this sense, we say the non-linear NNs are effectively linear over the domain of interest. We have shown this for MNIST successfully. From an application standpoint, mapping to dominant quasi-eigenvectors could be useful for medical imaging, diagnosis and prognosis if, \textit{e.g.}, the labels denoted the severity of a  disease; for predicting new materials if the labels denoted specific material features or external physical parameters.

\section*{Acknowledgements}
This research was supported in part by Lilly Endowment, Inc., through its support for the Indiana University Pervasive Technology Institute.

This work is partially supported by the Biocomplexity Institute at Indiana University, National Science Foundation grant 1720625 and National Institutes of Health grant NIGMS R01 GM122424.

\newpage


\bibliography{apssamp}

\end{document}